\documentclass[runningheads]{llncs}
\usepackage[utf8]{inputenc}
\usepackage[english]{babel}

\usepackage{makeidx}
\usepackage{graphicx}
\usepackage{amsmath,amssymb} %
\usepackage{mathtools}
\usepackage{bm}
\usepackage{color}
\usepackage{xspace}
\usepackage{microtype}
\usepackage{listings}

\usepackage{tikz}
\usetikzlibrary{external,positioning,shapes,fit,calc}
\usepackage{booktabs}
\usepackage{multirow}
\usepackage{subcaption}

\usepackage{varioref}
\usepackage[pagebackref=false,breaklinks=true,letterpaper=true,colorlinks,bookmarks=false]{hyperref}
\usepackage{cleveref}
\crefname{section}{Section}{Sections}
\crefname{figure}{Figure}{Figure}
\crefname{table}{Table}{Table}
\crefname{equation}{Equation}{Equation}

\newcommand{\onedot}{.\xspace}
\newcommand{\etal}[1]{#1~et~al\onedot}

\newcommand{\eg}{e.\,g.,\xspace}

\newcommand{\cf}{cf\onedot}
\newcommand{\ie}{i.\,e.,\xspace}

\renewcommand{\vec}[1]{\boldsymbol{\mathbf{#1}}}

\newcommand{\icdar}{\textsc{Icdar13}\xspace}
\newcommand{\cvl}{\textsc{Cvl}\xspace}

\newcommand{\map}{\ensuremath{\mathrm{mAP}}\xspace}

\definecolor{faublue}{RGB}{0,51,102}

\newcommand\review[1]{#1}
\newcommand\subreview[1]{#1}

\begin{document}
\pagestyle{headings}
\mainmatter

\def\GCPR15SubNumber{98}

\title{Offline Writer Identification Using Convolutional Neural Network
Activation Features}
\authorrunning{V.~Christlein et al.}
\titlerunning{Writer Identification Using CNN Activation Features}
\author{Vincent Christlein, David Bernecker, Andreas Maier, Elli Angelopoulou}
\institute{Pattern Recognition Lab, Friedrich-Alexander-Universität Erlangen-Nürnberg
	\email{\{firstname.lastname\}@fau.de}
}

\maketitle

\begin{abstract}
Convolutional neural networks (CNNs) have recently become the state-of-the-art
tool for large-scale image classification. In this work we propose the use of
activation features from CNNs as local descriptors for writer identification. A
global descriptor is then formed by means of GMM supervector encoding, which is
further improved by normalization with the KL-Kernel.  We evaluate our method on
two publicly available datasets: the ICDAR 2013 benchmark database and the CVL
dataset. While we perform comparably to the state of the art on CVL, our
proposed method yields about $0.21$ absolute improvement in terms of \map on the
challenging bilingual ICDAR dataset. 
\end{abstract}

\section{Introduction}
\label{sec:introduction}
In contrast to physiological biometric identifiers like fingerprints or iris
scans, handwriting can be seen as a behavioral
identifier~\cite{Zhu00BPI}. It is influenced by factors like schooling or
aging. 
Finding an individual writer in a large data corpus is formally defined as
\emph{writer identification}. Typical applications lie in the fields of forensics or
security. However, writer identification recently also raised interest in the analysis
of historical texts~\cite{Brink12WIU,Gilliam10SIM}. 

The task can be categorized into a) \emph{online} writer identification, for
which temporal information of the text formation can be used, and b)
\emph{offline} writer identification which relies solely on the handwritten
text. The latter can be further categorized into \emph{allograph}-based and
\emph{textural}-based methods~\cite{Bulacu07TIW}. Allograph-based methods rely
on local descriptors computed from small letter parts (allographs).
Subsequently, a global document descriptor is computed by means of statistics
using a pretrained
vocabulary~\cite{Christlein14WIA,Fiel13WIW,Gilliam10SIM,Jain14CLF,Siddiqi10TIW}.
In contrast, textural-based methods rely on global statistics computed
from the handwritten text, \eg the ink width or angle
distribution~\cite{Brink12WIU,Djeddi14EOT,He14DHR,Siddiqi10TIW,Newell14WIU}. Both
methods can be combined to form a stronger global
descriptor~\cite{Bulacu07TIW,Schomaker04AWI,Wu14OTI}. 

In this work we propose an allograph-based method for offline writer
identification. In contrast to expert-designed features like SIFT, we use activation
features learned by a convolutional neural network (CNN). \review{This has the
advantage of obtaining features guided by the data. In each additional CNN layer
the script is indirectly analyzed on a higher level of abstraction.}
CNNs have been widely used in image retrieval and object classification, and are
among the top contenders on challenges like the Pascal-VOC or
ImageNet~\cite{Krizhevsky12ICD}. However, to the best of our knowledge CNNs have
not been used for writer identification so far. A reason might be that typically
the training and test sets of current writer identification datasets are
disjoint making it impossible to train a CNN for classification. Thus, we propose to
use CNNs not for the classification task but to learn local activation
features. Subsequently, the local descriptors are encoded to form global
feature vectors by means of GMM supervector
encoding~\cite{Christlein14WIA}. We also propose to use the Kullback-Leibler
kernel, instead of the Hellinger kernel, on top of mean-only adapted GMM
parameters. 
We show that this combination of activation features and encoding
method performs at least as well as the current state of the art on two
public datasets \icdar and \cvl. %

\section{Related Work}
\label{sec:related-work}

Allograph-based methods rely on a dictionary trained from
local descriptors. This dictionary is subsequently used to collect statistics from the
local descriptors of the query document.  These statistics are then aggregated
to form the global descriptor that is used to classify the document. 
Jain and Doerman proposed the use of vector
quantization~\cite{Jain13WIU} as encoding method. More recent work concentrates
on using Fisher vectors for aggregation~\cite{Fiel13WIW,Jain14CLF}.
While Fiel and Sablatnig~\cite{Fiel13WIW} propose to use
solely SIFT descriptors as the local descriptor, Jain and
Doermann~\cite{Jain14CLF} suggest to fuse multiple Fisher vectors computed from different
descriptors. In contrast, we will rely on the findings of
\etal{Christlein}~\cite{Christlein14WIA}.  They showed that a very well known
approach in speaker recognition, namely GMM supervector encoding, performs
better than both Fisher vectors and VLAD encoding. 

CNNs have been widely used in the field of image classification and object
recognition. In the ImageNet Large Scale Visual Recognition Challenge for
example, CNNs are among the top contenders~\cite{Krizhevsky12ICD}. In
document analysis, CNNs have been used for word spotting by
\etal{Jaderberg}~\cite{Jaderberg14DFT}, and for handwritten text recognition by
\etal{Bluche}~\cite{Bluche13FEC}. However, to the best of our knowledge, they
have not been used in the context of writer identification.

Compared to regular feed forward neural networks, convolutional neural networks
have fewer parameters that need to be trained due to sharing the weights of their
filters across the whole input patch. This makes them easier to train, while
not sacrificing classification performance for a smaller sized network.
Instead of using a CNN for direct classification, one can choose to use a CNN to
extract local features by interpreting the activations of the last hidden layer
as the feature vector.
\etal{Bluche}~\cite{Bluche13FEC} propose to use features learned by a CNN for
word recognition in conjunction with HMMs, and show that the learned features
outperform previous representations.  \etal{Gong}~\cite{Gong14MSO} employ a
similar approach for image classification. Their local activation features are
computed by calculating the activation of a pretrained CNN on the image itself,
and on patches of various scales extracted from the image.  The activations for
each scale are then aggregated using VLAD encoding. The final image
descriptor is formed by concatenating the resulting feature vectors from each
scale. 

\section{Writer Identification Pipeline}
\label{sec:pipeline}

\begin{figure}[t]
	\centering
\includegraphics[width=0.9\textwidth]{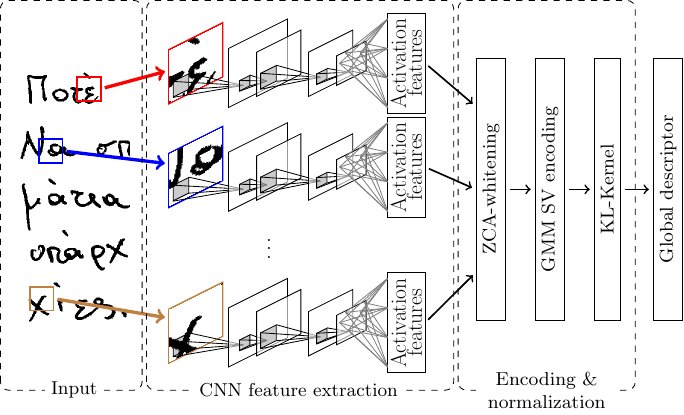}

	\caption{Overview of the encoding process. The two main steps are the feature
	extraction using a pretrained CNN, and the encoding step, where the local
features are agreggated using a pretrained GMM.}
	\label{fig:pipeline}
\end{figure}

Our proposed pipeline (\cf \cref{fig:pipeline}) consists of three main steps: the
feature extraction from image patches using a CNN; the aggregation of all the
local features from one document into one global descriptor; and the successive
normalization of this descriptor. A pretrained CNN and a pretrained GMM are
required for feature extraction and encoding, respectively. 

\subsection{Convolutional Neural Networks}

\begin{figure}[t]
	\centering
\includegraphics[width=0.8\textwidth]{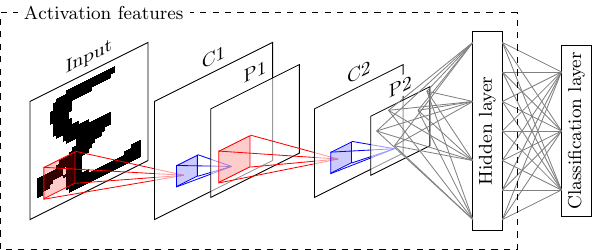}
\caption{Schematic representation of the used CNN. C1 and C2 are convolutional
	layers (red connections). P1 and P2 are max pooling layers (blue connections).
	The last three layers are fully connected (gray connections).  After training
	only the part of the net inside the dashed box (activation features) is kept.
	The activations of the hidden layer become the local descriptor for the image
patch.} \label{fig:cnn}
\end{figure}
In our pipeline the CNN is only used to calculate a feature representation of a
small image patch, but not for directly identifying the writer. The training of
the CNN, however, has to be performed by backpropagation, which requires labels
for the individual patches. Therefore, during the training phase, the last layer
of our network consists of 100 SoftMax nodes, representing the writer IDs of the
\icdar training set. After the training, this last layer is discarded and
the remaining layers are used to generate the feature representation for the
image patches. The architecture of the CNN we use is shown in \cref{fig:cnn},
where the dashed box marks the part of the CNN that is kept after the training
procedure.

The CNN consists of 6 layers in total. The first layer is a convolutional layer,
followed by a pooling layer. In the convolutional layer, the input patch is
convolved with $16$ filters. The pooling layer is then used to reduce the
dimensions of the filter responses by performing a max pooling over regions of
size $2\times2$ or $3\times3$.  The two subsequent layers follow the same
principle: a convolutional layer with $256$ filters is followed by a pooling
layer.  These first four layers constitute the convolutional part of the
network. The output of the second pooling layer is next transformed into a 1-D
vector which is fed into a layer of hidden nodes. For all of these layers
rectified linear units (ReLU) are used as nodes.  The last layer then consists
of 100 nodes with a SoftMax activation function. They are used for
classification during the training.

The training set consists of patches extracted from the \icdar training set
that are centered on the contour of the writing. For each of the $100$ writers,
\icdar contains four images, two of Greek handwritten text and
two of English handwritten text. We further divided this
set into a training and test set, by using patches from the first
English and Greek text for training, and patches from the second English and
Greek text for testing the trained convolutional network. The training and test set
consist independently of 4 million image patches of size $32\times32$. \review{The image
patches are not preprocessed in any manner.}

The training is performed by using the CUDA capabilities of the neural network
library Torch~\cite{Collobert11TAM}. All the CNNs are trained using the Torch implementation of
stochastic gradient descent (SGD) with a learning rate of $0.01$ for 20 epochs.
For the first five epochs of training a Nesterov momentum $m=0.9$ is used to
speed up the training process.

\subsection{GMM Supervector Encoding}
Given the local activation features, we need to aggregate them to form one
global descriptor for each document. For this task we use a variant of the GMM
supervector approach of \etal{Christlein}~\cite{Christlein14WIA}. 

In the training step a Gaussian mixture model (GMM) is
trained as the dictionary from a set of \review{ZCA}-whitened activation features. This
dictionary is subsequently used to encode the local descriptors by calculating
their statistics with regard to the dictionary.  The $K$-component GMM is
denoted by $\lambda = \lbrace w_k,\, \vec{\mu}_k,\, \vec{\Sigma}_k\,\vert
k=1,\ldots,K\rbrace$, where $w_k$, $\vec{\mu}_k$ and $\vec{\Sigma}_k$ are the
mixture weight, mean vector and diagonal covariance matrix for mixture $k$,
respectively. The parameters $\lambda$ are estimated with the
expectation-maximization (EM) algorithm~\cite{Dempster77MLF}.

Given the pretrained GMM and one document, the parameters $\lambda$ are first
adapted to all activation features extracted from the document by means of a
maximum-a-posteriori (MAP) step. Using a data-dependent mixing coefficient they
are coupled with the parameters of the pretrained GMM. This leads to different
mixtures being adapted depending on the current set of activation
features~\cite{Reynolds00SVU}. Given the descriptors $\vec{X} = \lbrace
\vec{x}_t, \vec{x}_t \in \mathcal{R}^D, t = 1,\ldots T\rbrace$  of a document,
first the posterior probabilities $\gamma_t(k)$ for each $\vec{x}_t$ and
Gaussian mixture $g_k(\vec{x})$ are computed as:
\begin{equation}
	\gamma_t(k) =\frac{w_kg_k(\vec{x}_t)}{\sum_{j=1}^Kw_jg_j(\vec{x}_t)}\,. 
\end{equation}

\review{Since the covariances and weights give only a slight improvement in
	accuracy~\cite{Christlein14WIA},  we chose to adapt only the means of
the mixtures, thus, reducing the size of the output supervector and lowering the
computational effort}. The first order statistics are computed as:
\begin{equation}
	\hat{\vec{\mu}}_k = \frac{1}{n_k}	\sum_{i=1}^T\gamma_t(k)\vec{x}_t\,,
\end{equation}
where $n_k=\sum_{t=1}^T \gamma_t(k)$. Then, these new means are mixed with the
original GMM means:
\begin{equation}
	\tilde{\vec{\mu}}_k = \alpha_k \hat{\vec{\mu}}_k + (1-\alpha_k) \vec{\mu}_k
	\,,
\end{equation}
where $\alpha_k$ denotes a data dependent adaptation coefficient. It is computed by
	$\alpha_k = \frac{n_k}{n_k + \tau}$, 
where $\tau$ is a relevance factor. The new parameters of the mixed
GMM are then concatenated forming the 
GMM supervector:
	$\vec{s} = \big( \tilde{\vec{\mu}}_1^{\top}, \ldots,
	\tilde{\vec{\mu}}_K^{\top} \big)^{\top}$. %
This global descriptor $\vec{s}$ is a $KD$ dimensional vector which is
eventually used for nearest neighbor search using the cosine-distance as metric. 

\subsection{Normalization}
While contrast-normalization is an often used intermediate step in CNN
training~\cite{Bengio11UAT}, we employ \review{ZCA} whitening to decorrelate the
activation features followed by a global $L_2$ normalization.  We will show that
the accuracy of the GMM supervector benefits greatly from this normalization
step.

Additionally, our GMM supervector is normalized, too. \etal{Christlein}
suggested to normalize the full GMM supervector (consisting of the adapted
weight, mean and covariance parameters) using power normalization with a power
of $0.5$ prior to a $L_2$ normalization~\cite{Christlein14WIA}.  Effectively
this results in applying the Hellinger kernel. In contrast, we employ a kernel
derived from the symmetrized Kullback-Leibler divergence~\cite{Xu10EHG} to
normalize the adapted components:
\begin{equation}
	\mathring{\vec{\mu}}_k
	= \sqrt{w_k} \vec{\sigma}^{-\frac{1}{2}}_k
		\tilde{\vec{\mu}}_k \,, 
\end{equation}
where $\vec{\sigma}_k$ is the vector of the diagonal elements of the covariance
matrix $\vec{\Sigma}$ of the trained Gaussian mixture $k$. This implicitly
encodes information contained in the variances and weights of the GMM, although
only the means were adapted in the main encoding step. The normalized
supervector becomes $\mathring{\vec{s}} = \big(
\mathring{\vec{\mu}}_1^{\top}, \ldots, \mathring{\vec{\mu}}_K^{\top} \big)
^{\top}$.

\subsection{Implementation Notes}
For the computation of the posteriors, we set all but the ten highest posterior
probabilities computed from each descriptor to zero. Consequently, we compute
the adaptation only for the data having non-zero posteriors. This has the effect
of reducing the computational cost with nearly no loss in accuracy. Similar to
the work of \etal{Christlein}~\cite{Christlein14WIA}, we used $100$ Gaussian
mixtures, but raised the relevance factor $\tau$ to $68$ which was found to
slightly improve the results.

\section{Evaluation}
\label{sec:evaluation}

\subsection{Datasets}

We use two different datasets for evaluation: the \icdar benchmark
set~\cite{Louloudis13ICO}
and the \cvl dataset~\cite{Kleber13CDA}. Both are publicly available and have
been used in many recent
publications~\cite{Christlein14WIA,Fiel13WIW,Jain14CLF}.

\subsubsection{ICDAR13~\cite{Louloudis13ICO}} The \icdar benchmark set
is separated into a training set consisting of documents from 100 writers and a
writer independent test
set consisting of documents from 250 writers.
Each writer contributed four documents. Two
are written in Greek, and two are written in English. This provides for a
challenging cross-language writer identification.

\subsubsection{CVL~\cite{Kleber13CDA}} The \cvl dataset consists
of 310 writers. The dataset is split in a training set and a test set without
overlap of the writers. The training set contains 27 writers contributing seven
documents each. The test set consists of 283 writers who contributed five
documents each.  One document out of the five (seven) documents is written in
German, the others in English. Note that we binarized the documents using
Otsu's method.

\subsection{Metrics}

To evaluate our experiments we use the mean average precision (\map) and the
hard TOP-$k$ scores. Both are common metrics in information retrieval tasks.
Given a query document from one writer, an ordered list of documents is
returned, where the first returned document is regarded as being the closest to
the query document.  The \map then is the mean of the average precision
($\operatorname{aP}$) over all queries. $\operatorname{aP}$ is defined as
\begin{equation}
	\operatorname{aP} =
	\frac{\sum_{k=1}^nP(k)\cdot\operatorname{rel}(k)}{\text{\#relevant
	documents}}\,.
\end{equation}
Given the ordered list of documents for a query document, the
$\operatorname{aP}$ averages over $P(k)$, the precision at rank $k$, that is
given by the number of documents from the same writer in the query up to rank
$k$ divided by $k$.  $\operatorname{rel}(k)$ is an indicator function that is
one if the document retrieved at rank $k$ is from the same writer and zero
otherwise. 

The hard TOP-$k$ scores are determined by calculating the percentage of queries,
where the $k$ highest ranked documents were from the same writer\subreview{, \eg
the hard TOP-$3$ denotes the probability that the three best ranked documents stem from
the correct writer.}

\subsection{Convolutional Neural Network Parameters}
\begin{table}[t]
\caption{Evaluation of different CNN configurations on the \icdar
	training set} 
	\label{tab:params}
\hfill
\begin{subtable}[t]{0.8\textwidth}
	\centering
  \begin{tabular}{ccccc}
		\toprule
		Filter configuration & C1 & P1 & C2 & P2\\
		\midrule
		\textbf{A} & $5\times5$\, & \,$2\times2$\, & \,$5\times5$\, & \,$2\times2$ \\
		\textbf{B} & $7\times7$\, & \,$2\times2$\, & \,$5\times5$\, & \,$3\times3$ \\
	\bottomrule
\end{tabular}
  \caption{Convolutional and pooling layer configurations of the CNN}
  \label{tab:filters}
\end{subtable}
\hspace*{\fill}\\
	\hfill
	\begin{subtable}[t]{0.54\textwidth}
\centering
\begin{tabular}{cccc}
	\toprule 
	\multirow{2}{40pt}{Filter size} & \multicolumn{3}{c}{No.~hidden nodes}\\
	& 64 & 128 & 256\\
	\midrule
	\textbf{A} & 38.2/23.2 & 49.3/23.7 & 55.0/24.5 \\
	\textbf{B} & 40.3/21.0 & 45.6/22.4 & 53.5/23.0 \\
	\bottomrule
\end{tabular}
\caption{Classification accuracy [\%] using the classification layer of the CNN
(train/test)}
\label{tab:paramsa}
\end{subtable}
\hfill
\begin{subtable}[t]{0.44\textwidth}
\centering
\begin{tabular}{cccc}
	\toprule 
	\multirow{2}{40pt}{Filter size} & \multicolumn{3}{c}{No.~hidden nodes}\\
	& 64 & 128 & 256\\
	\midrule
	\textbf{A} & $0.937$\, & \,$0.926$\, & \,$0.895$ \\
	\textbf{B} & $0.948$\, & \,$0.929$\, & \,$0.910$ \\
	\bottomrule
\end{tabular}
\caption{Averaged \map of VLAD encoding}
\label{tab:paramsb}
\end{subtable}
\end{table}

With the CNN architecture fixed to two convolutional and one hidden layer there
are two main parameters that are essential for the performance of the trained
activation features: the filter size, and the number of hidden nodes in the last
layer, \ie the size of the output descriptor. We conducted some preliminary
experiments using the \icdar training set to determine the optimal parameters
for the chosen network architecture. We evaluated two different setups of the
filter and pooling sizes for the convolutional layers. The values for the two
configurations \textbf{A} and \textbf{B} are shown in \cref{tab:filters}.
Comparing the two configurations shows that, \textbf{B} uses larger filters and
pooling sizes and should therefore be more insensitive to translations of the
patches. For both filter sizes we also evaluated the effect of the output
feature size by using three different numbers of hidden nodes in the last layer:
$64$, $128$, and $256$.

For these preliminary experiments we used VLAD encoding~\cite{Jegou12ALI}
instead of GMM supervectors due to its faster computation time. VLAD is a
non-probabilistic version of Fisher vectors which hard-encodes the first order
statistics, \ie
$	\vec{s}_k = \sum_{\vec{x}_t \in \tilde{\vec{X}}} (\vec{x}_t -\vec{\mu}_k)$, 
where $\tilde{\vec{X}}$ refers to the set of descriptors for which the cluster
center $\vec{\mu}_k$ is the closest one. The dictionary can be efficiently
computed by using a mini-batch version of $k$-means~\cite{Sculley10WSK}. 
We report the average \map over the results of 10 VLAD-encoding runs.

Besides the network configurations, \cref{tab:params} shows the classification
accuracy obtained with the CNN including the classification layer on the test
set after 20 epochs of training in part (b) and the averaged \map of 10 runs of
VLAD encoding in part (c).  Interestingly, the results for both evaluation
approaches are almost complementary. The CNN alone reaches the best results for
smaller filters and a large number of hidden nodes, while the VLAD encoding
prefers larger filters and a smaller size of the activation features vector (\ie
number of hidden nodes). A possible explanation might be that, for a larger
number of hidden nodes the activations of the hidden layer are less descriptive
for discerning between writers because the connections between the hidden and
the classification layer take over that part. In contrast, for a small number of
hidden nodes, the descriptiveness of the activations of the hidden layer seems
to be higher, making them more suitable for use as features independent from the
classification layer of the CNN. It should also be noted that the classification
accuracy of the CNN is already quite impressive considering that the
classification is performed using only a single patch of size $32\times32$ for
$100$ different writers/classes.
Since configuration \textbf{B} shows the highest \map, this configuration of the
CNN is used for all of the following experiments.

\subsection{Performance Analysis}
\begin{table}[t]
	\centering
	\caption{The influence of different parts of the pipeline on the
	\icdar test set}
	\label{tab:comparison1}
	\begin{subtable}{0.45\textwidth}
		\centering
		\begin{tabular}{llc}
			\toprule 
			\multicolumn{2}{l}{Method} & \map\\
			\midrule
			RootSIFT &+ $\text{SV}_{\text{wmc,ssr+l2}}$~\cite{Christlein14WIA}  & 0.671\\
			RootSIFT &+ $\text{SV}_{\text{m,kl}}$ & 0.680\\
			SURF &+ $\text{SV}_{\text{m,kl}}$ & 0.718\\
			$\text{CNN-AF}$ &+ $\text{SV}_{\text{m,kl}}$ &  0.860\\
			\bottomrule
		\end{tabular}
	\caption{Comparison of different local descriptors}
	\label{tab:diff_desc}
	\end{subtable}
	\hfill
	\begin{subtable}{0.45\textwidth}
		\centering
		\begin{tabular}{lc}
			\toprule
			Method & \map\\
			\midrule
			$\text{CNN-AF}_\text{pwh} + \text{SV}_{\text{m,kl}}$ & 0.880\\
			$\text{CNN-AF}_\text{zwh} + \text{SV}_{\text{m,kl}}$ & \review{0.886}\\
			$\text{CNN-AF}_\text{zwh} + \text{SV}_{\text{wmc,ssr+l2}}$ & \review{0.877}\\
			$\text{CNN-AF}_\text{zwh} + \text{FV}$ & \review{0.866}\\
			\bottomrule
		\end{tabular}
		\caption{\review{Influence of different whitening and encoding methods}}
		\label{tab:diff_enc}
	\end{subtable}
	\hspace*{\fill}
\end{table}

We now investigate the influence of the individual steps in our pipeline.  We
replace the CNN activation features by other local descriptors. We also examine
the influence of applying \review{ZCA- and PCA}-whitening to the CNN activation features. Lastly,
we evaluate the replacement of the GMM supervectors with other encoding methods.

\cref{tab:diff_desc} compares the learned activation features with SURF and
RootSIFT. Both have been used successfully for offline writer identification by
Jain and Doermann~\cite{Jain14CLF} and \etal{Christlein}~\cite{Christlein14WIA},
respectively. Interestingly, SURF performs better than RootSIFT. However, our
proposed activation features outperform both descriptors by $0.14$ and $0.18$
\map, respectively.

\cref{tab:diff_enc} shows the effect of decorrelating the activation features
using \review{PCA and \review{ZCA} whitening ($\text{CNN-AF}_\text{pwh} + \text{SV}_{\text{m,kl}}$ vs.
$\text{CNN-AF}_\text{zwh} + \text{SV}_{\text{m,kl}}$)} and the comparison with the
other encoding methods. $\text{CNN-AF}_\text{zwh} +
\text{SV}_{\text{wmc,ssr+l2}}$ is using GMM supervectors as proposed by
\etal{Christlein}~\cite{Christlein14WIA} and $\text{CNN-AF}_\text{zwh} +
\text{FV}$ uses Fisher vectors as proposed by
\etal{Sanchez}~\cite{Sanchez13ICW}. The SV encoding by \etal{Christlein} adapts
all components (weights, means, covariances) while the FV encoding uses the
means and covariances.  Both methods use power normalization (power of $0.5$)
followed by $l_2$ normalization instead of the KL-kernel normalization.

The decorrelation of the features brings an improvement of $0.02$ \map,
\review{with ZCA giving slightly better results than PCA}. The decorrelated score with
the proposed method also outperforms the two other encoding methods.

\subsection{Comparison with the State of the Art}

\cref{tab:icdar_results} and \cref{tab:cvl_results} show the results achieved
with the complete pipeline on the \icdar and \cvl test sets, respectively.
\review{We compare with the state of the art
\footnote{\review{The methods~\cite{Jain14CLF} and~\cite{He14DHR} did not
provide results on the full \icdar dataset.}} and SURF descriptors encoded with
	GMM supervectors, \cf~\cref{tab:diff_desc}}. Since
the \cvl training set is too small to compute a comparable GMM, we used the GMM
and \review{ZCA} transformation matrix estimated on the \icdar training set for
evaluating the pipeline on the \cvl dataset. On both datasets the proposed
pipeline using CNN activation features outperforms the previous methods in terms
of \map. The increase in performance is particularly evident on the
complete \icdar test set, where our method achieves an absolute improvement of
0.21 \map. \review{This is significantly better than the state of the
art~\cite{Christlein14WIA} (permutation test: $p \ll
0.05$). On the \cvl dataset we achieve comparable results to the state of the
art (permutation test: $p=0.11$).	
However note that a) the \icdar dataset is much more
	challenging due to its bilingual nature, and b) that we have not 
	trained explicitly for the CVL dataset. Thus, our results show that the
	features learned from the ICDAR training set can generally be used for other
datasets, too.}
We believe that the results could be further improved if the \cvl training set would
be incorporated into the training of the CNN activation features.

\cref{tab:icdar_lang} shows the results for evaluating the
Greek and English subsets of the \icdar test set independently. Again, the proposed method further
improves the already high scores of the previous methods.
\begin{table}[t]
	\caption{Hard criterion TOP-$k$ scores and \map evaluated on \icdar (test
	set)}
	\label{tab:all_results}
	\hfill
	\begin{subtable}[t]{0.46\textwidth}
	\centering
\begin{tabular}{lcccc}
	\toprule
	\phantom{Gg} & \multirow{2}{28pt}{TOP-$1$} & \multirow{2}{28pt}{TOP-$2$} &
	\multirow{2}{28pt}{TOP-$3$} & \multirow{2}{28pt}{\map}\\
	& & & & \\
	\midrule
	CS~\cite{Jain13WIU} & 0.951 & 0.196 & 0.071 & NA\\ 
	SV~\cite{Christlein14WIA} & 0.971 & 0.428 & 0.238 & 0.671\\
	\review{SURF} & \review{0.967} & \review{0.551} & \review{0.273} &
	\review{0.718}\\
  \review{Proposed} & \review{\textbf{0.989}} & \review{\textbf{0.832}} &
	\review{\textbf{0.613}} & \review{\textbf{0.886}} \\
	\bottomrule
\end{tabular}
\caption{Complete \icdar test set}
\label{tab:icdar_results}
\end{subtable}
\hfill
\begin{subtable}[t]{0.48\textwidth}
	\centering
	\begin{tabular}{lcc|cc}
		\toprule
		& \multicolumn{2}{c|}{Greek} & \multicolumn{2}{c}{English}\\
		& TOP-$1$ & \map & TOP-$1$ & \map\\
		\midrule
		$\Delta$-n H.~\cite{He14DHR} & 0.960 & NA & 0.934 & NA\\
		Comb.~\cite{Jain14CLF} & 0.992 & 0.995 & 0.974 & 0.979\\
		\review{SURF} & \review{0.950} & \review{0.965} & \review{0.956} &
		\review{0.964}\\
		Proposed & \textbf{0.996} & \textbf{0.998}& \textbf{0.976} & \textbf{0.981}\\
		\bottomrule
	\end{tabular}
	\caption{\icdar language subsets}
	\label{tab:icdar_lang}
	\end{subtable}
\end{table}
\begin{table}[t]
\centering
\caption{Hard criterion and \map evaluated on \cvl}
\label{tab:cvl_results}
\begin{tabular}{lccccc}
	\toprule
				& TOP-$1$ & TOP-$2$ & TOP-$3$ & TOP-$4$ & \map\\
	\midrule
	FV~\cite{Fiel13WIW} & 0.978 & 0.956 & 0.894 & 0.758 & NA \\
	Comb~\cite{Jain14CLF} & \textbf{0.994} & 0.983 & 0.948 & 0.829 & 0.969\\
	SV~\cite{Christlein14WIA} & 0.992 & 0.981 & 0.958 & 0.887 & 0.971\\
SURF & 0.986 &  0.973 & 0.948 & 0.836 & 0.958\\
	\review{Proposed} & \review{\textbf{0.994}} & \review{\textbf{0.988}} &
	\review{\textbf{0.973}} & \review{\textbf{0.926}} &	\review{\textbf{0.978}}\\
\bottomrule
\end{tabular}
\end{table}

\section{Conclusion}
\label{sec:conclusion}

The writer identification method proposed in this paper exploits activation
features learned by a deep CNN, which in comparison to traditional local
descriptors like SIFT or SURF yield higher \map scores on the \icdar and \cvl
datasets. On the \icdar test set, an increase of
about 0.21 \map is achieved with this new set of features. We show in our
experiments that the retrieval rate is strongly influenced by the design choices
of the CNN architecture.  
The local activation features are encoded using a modified variant of the GMM
supervectors approach.  However, we adapt only the means of the Gaussian
mixtures in the aggregation step. Subsequently, the supervector is normalized
using the KL-kernel.  By implicitly adding information contained in the weights
and covariances of the mixtures in the normalization step, the performance is
increased %
while at the same time halving the dimensionality of the global descriptor.

For future work, we would like to explore larger and more complex CNN
architectures and recent discoveries like the benefit of
$L_p$-pooling~\cite{Sermanet12CNN} instead of max pooling and normalization of
activations after convolutional layers of the network. There is also still room
for improvement in the encoding step of the local descriptors, where democratic
aggregation~\cite{Jegou14TEA} or higher order VLAD~\cite{Peng14BVW} could
further improve the writer identification rates.

\section*{Acknowledgments}
This work has been supported by the German Federal Ministry of Education
and Research (BMBF), grant-nr. 01UG1236a. The contents of this 
publication are the sole responsibility of the authors.

\clearpage
\bibliographystyle{splncs03}
\bibliography{wacv13-icdar15,wacv13-icdar15b}

\begin{thebibliography}{10}
\providecommand{\url}[1]{\texttt{#1}}
\providecommand{\urlprefix}{URL }

\bibitem{Bengio11UAT}
Bengio, Y.: {Deep Learning of Representations for Unsupervised and Transfer
  Learning}. In: Unsupervised and Transfer Learning, Challenges in Machine
  Learning. vol.~7, pp. 19--41. Bellevue (Jun 2011)

\bibitem{Bluche13FEC}
Bluche, T., Ney, H., Kermorvant, C.: {Feature Extraction with Convolutional
  Neural Networks for Handwritten Word Recognition}. In: 2013 12th
  International Conference on Document Analysis and Recognition. pp. 285--289.
  Buffalo (Aug 2013)

\bibitem{Brink12WIU}
Brink, A., Smit, J., Bulacu, M., Schomaker, L.: {Writer Identification Using
  Directional Ink-Trace Width Measurements}. Pattern Recognition  45(1),
  162--171 (Jan 2012)

\bibitem{Bulacu07TIW}
Bulacu, M., Schomaker, L.: {Text-Independent Writer Identification and
  Verification Using Textural and Allographic Features}. Pattern Analysis and
  Machine Intelligence, IEEE Transactions on  29(4),  701--17 (Apr 2007)

\bibitem{Christlein14WIA}
Christlein, V., Bernecker, D., Honig, F., Angelopoulou, E.: {Writer
  Identification and Verification Using GMM Supervectors}. In: Applications of
  Computer Vision (WACV), 2014 IEEE Winter Conference on. pp. 998--1005 (Mar
  2014)

\bibitem{Collobert11TAM}
Collobert, R., Kavukcuoglu, K., Farabet, C.: {Torch7: A Matlab-like Environment
  for Machine Learning}. In: Big Learning, Workshop on Advances in Neural
  Information Processing Systems 24 (NIPS 2011). Granada (Dec 2011)

\bibitem{Dempster77MLF}
Dempster, A., Laird, N., Rubin, D.: {Maximum Likelihood from Incomplete Data
  via the EM Algorithm}. Journal of the Royal Statistical Society. Series B
  (Methodological)  39(1),  1--38 (1977)

\bibitem{Djeddi14EOT}
Djeddi, C., Meslati, L.S., Siddiqi, I., Ennaji, A., Abed, H.E., Gattal, A.:
  {Evaluation of Texture Features for Offline Arabic Writer Identification}.
  In: Document Analysis Systems (DAS), 2014 11th IAPR International Workshop
  on. pp. 8--12. Tours (Apr 2014)

\bibitem{Fiel13WIW}
Fiel, S., Sablatnig, R.: {Writer Identification and Writer Retrieval using the
  Fisher Vector on Visual Vocabularies}. In: Document Analysis and Recognition
  (ICDAR), 2013 12th International Conference on. pp. 545--549. Washington DC
  (Aug 2013)

\bibitem{Gilliam10SIM}
Gilliam, T., Wilson, R., Clark, J.: {Scribe Identification in Medieval English
  Manuscripts}. In: Pattern Recognition (ICPR), 2010 20th International
  Conference on. pp. 1880--1883. Istanbul (Aug 2010)

\bibitem{Gong14MSO}
Gong, Y., Wang, L., Guo, R., Lazebnik, S.: {Multi-scale Orderless Pooling of
  Deep Convolutional Activation Features}. In: Fleet, D., Pajdla, T., Schiele,
  B., Tuytelaars, T. (eds.) Computer Vision – ECCV 2014, vol. 8695, pp.
  392--407. Springer International Publishing, Zurich (Sep 2014)

\bibitem{He14DHR}
He, S., Schomaker, L.: {Delta-n Hinge: Rotation-Invariant Features for Writer
  Identification}. In: Pattern Recognition (ICPR), 2014 22nd International
  Conference on. pp. 2023--2028. Stockholm (Aug 2014)

\bibitem{Jaderberg14DFT}
Jaderberg, M., Vedaldi, A., Zisserman, A.: {Deep Features for Text Spotting}.
  In: Computer Vision – ECCV 2014, vol. 8692, pp. 512--528. Springer
  International Publishing, Zurich (Sep 2014)

\bibitem{Jain13WIU}
Jain, R., Doermann, D.: {Writer Identification Using an Alphabet of Contour
  Gradient Descriptors}. In: Document Analysis and Recognition (ICDAR),
  International Conference on. pp. 550--554. Buffalo (Aug 2013)

\bibitem{Jain14CLF}
Jain, R., Doermann, D.: {Combining Local Features for Offline Writer
  Identification}. In: Frontiers in Handwriting Recognition (ICFHR), 2014 14th
  International Conference on. pp. 583--588. Heraklion (Sep 2014)

\bibitem{Jegou14TEA}
J\'{e}gou, H., Zisserman, A.: {Triangulation Embedding and Democratic
  Aggregation for Image Search}. In: Computer Vision and Pattern Recognition
  (CVPR), 2014 IEEE Conference on. pp. 3310--3317. Columbus (Jun 2014)

\bibitem{Jegou12ALI}
J\'{e}gou, H., Perronnin, F., Douze, M., S\'{a}nchez, J., P\'{e}rez, P.,
  Schmid, C.: {Aggregating Local Image Descriptors into Compact Codes}. Pattern
  Analysis and Machine Intelligence, IEEE Transactions on  34(9),  1704--1716
  (Sep 2012)

\bibitem{Kleber13CDA}
Kleber, F., Fiel, S., Diem, M., Sablatnig, R.: {CVL-DataBase: An Off-Line
  Database for Writer Retrieval, Writer Identification and Word Spotting}. In:
  Document Analysis and Recognition (ICDAR), 2013 12th International Conference
  on. pp. 560 -- 564. Washington DC (Aug 2013)

\bibitem{Krizhevsky12ICD}
Krizhevsky, A., Sutskever, I., Hinton, G.E.: {ImageNet Classification with Deep
  Convolutional Neural Networks}. In: Advances In Neural Information Processing
  Systems 25, pp. 1097----1105. Curran Associates, Inc. (2012)

\bibitem{Louloudis13ICO}
Louloudis, G., Gatos, B., Stamatopoulos, N., Papandreou, A.: {ICDAR 2013
  Competition on Writer Identification}. In: Document Analysis and Recognition
  (ICDAR), 2013 12th International Conference on. pp. 1397--1401. Washington DC
  (Aug 2013)

\bibitem{Newell14WIU}
Newell, A.J.A., Griffin, L.D.L.: {Writer Identification Using Oriented Basic
  Image Features and the Delta Encoding}. Pattern Recognition  47(6),
  2255--2265 (Jun 2014)

\bibitem{Peng14BVW}
Peng, X., Wang, L., Qiao, Y., Peng, Q.: {Boosting VLAD with Supervised
  Dictionary Learning and High-Order Statistics}. In: Fleet, D., Pajdla, T.,
  Schiele, B., Tuytelaars, T. (eds.) Computer Vision – ECCV 2014, Lecture
  Notes in Computer Science, vol. 8691, pp. 660--674. Springer International
  Publishing, Zurich (Sep 2014)

\bibitem{Reynolds00SVU}
Reynolds, D.A., Quatieri, T.F., Dunn, R.B.: {Speaker Verification Using Adapted
  Gaussian Mixture Models}. Digital Signal Processing  10(1-3),  19--41 (2000)

\bibitem{Sanchez13ICW}
S\'{a}nchez, J., Perronnin, F., Mensink, T., Verbeek, J.: {Image Classification
  with the Fisher Vector: Theory and Practice}. International Journal of
  Computer Vision  105(3),  222--245 (2013)

\bibitem{Schomaker04AWI}
Schomaker, L., Bulacu, M.: {Automatic Writer Identification Using
  Connected-Component Contours and Edge-Based Features of Uppercase Western
  Script}. Pattern Analysis and Machine Intelligence, IEEE Transactions on
  26(6),  787--798 (2004)

\bibitem{Sculley10WSK}
Sculley, D.: {Web-scale K-means Clustering}. In: World Wide Web, 19th
  International Conference on. pp. 1177--1178. WWW '10, ACM, New York (Apr
  2010)

\bibitem{Sermanet12CNN}
Sermanet, P., Chintala, S., LeCun, Y.: {Convolutional Neural Networks Applied
  to House Numbers Digit Classification}. In: Pattern Recognition (ICPR), 2012
  21st International Conference on. pp. 3288--3291. IEEE, Tsukuba (Nov 2012)

\bibitem{Siddiqi10TIW}
Siddiqi, I., Vincent, N.: {Text Independent Writer Recognition using Redundant
  Writing Patterns with Contour-Based Orientation and Curvature Features}.
  Pattern Recognition  43(11),  3853--3865 (2010)

\bibitem{Wu14OTI}
Wu, X., Tang, Y., Bu, W.: {Offline Text-Independent Writer Identification Based
  on Scale Invariant Feature Transform}. Information Forensics and Security,
  IEEE Transactions on  9(3),  526--536 (Mar 2014)

\bibitem{Xu10EHG}
Xu, M., Zhou, X., Li, Z., Dai, B., Huang, T.S.: {Extended Hierarchical
  Gaussianization for Scene Classification}. In: Image Processing (ICIP), 2010
  17th IEEE International Conference on. pp. 1837--1840. Hong Kong (Sep 2010)

\bibitem{Zhu00BPI}
Zhu, Y.Z.Y., Tan, T.T.T., Wang, Y.W.Y.: {Biometric Personal Identification
  Based on Handwriting}. In: 15th International Conference on Pattern
  Recognition (ICPR). vol.~2, pp. 2--5. Barcelona (Sep 2000)

\end{thebibliography}

\end{document}